\documentclass{article} 
\usepackage{iclr2027_conference,times}

\usepackage{amsmath,amsfonts,bm}

\def\eqref#1{equation~\ref{#1}}

\def\1{\bm{1}}

\def\mA{{\bm{A}}}
\def\mB{{\bm{B}}}

\def\mD{{\bm{D}}}

\def\mG{{\bm{G}}}

\def\mI{{\bm{I}}}

\def\mP{{\bm{P}}}

\def\mU{{\bm{U}}}
\def\mV{{\bm{V}}}
\def\mW{{\bm{W}}}

\def\mZ{{\bm{Z}}}

\def\mSigma{{\bm{\Sigma}}}

\DeclareMathAlphabet{\mathsfit}{\encodingdefault}{\sfdefault}{m}{sl}
\SetMathAlphabet{\mathsfit}{bold}{\encodingdefault}{\sfdefault}{bx}{n}

\def\gC{{\mathcal{C}}}
\def\gD{{\mathcal{D}}}

\usepackage{wrapfig}
\usepackage{needspace}
\usepackage{amsthm}
\newtheorem{theorem}{Theorem}
\newtheorem{proposition}[theorem]{Proposition}
\usepackage{hyperref}
\usepackage{url}
\usepackage{booktabs}
\usepackage{graphicx}
\usepackage{multirow}
\usepackage{placeins}
\usepackage{amsmath}
\usepackage{enumitem}
\usepackage{algorithm}
\usepackage{algpseudocode}
\usepackage{xcolor,colortbl}
\usepackage{float}
\usepackage{capt-of}
\usepackage{amssymb}
\usepackage{svg}
\usepackage[most]{tcolorbox}
\newtcolorbox{remarkbox}{
  colback=gray!8,
  colframe=gray!8,
  boxrule=0pt,
  arc=0pt,
  outer arc=0pt,
  boxsep=0pt,
  left=4pt,
  right=4pt,
  top=4pt,
  bottom=4pt,
  before skip=6pt,
  after skip=6pt,
  breakable
}
\let\OriginalAddContentsLine\addcontentsline

\title{ReGRA: Residual-Gradient Adaptation for Complementary Low-Rank Fine-Tuning}

\author{Fanyao Li\thanks{Equal contribution.},\; Yuxi Liu\footnotemark[1],\; Xiaohao Liu,\; He Chang,\; Zhaoxin Wang,\; Zhulin Tao\thanks{Corresponding author.} \\
Communication University of China, National University of Singapore, Xidian University \\
\texttt{fanyao.li@mails.cuc.edu.cn, liuyuxi@cuc.edu.cn} \\
\texttt{xiaohao.liu@u.nus.edu, hechangcuc@cuc.edu.cn} \\
\texttt{wangzhaoxin@stu.xidian.edu.cn, taozhulin@gmail.com}
}

\iclrfinalcopy 
\begin{document}

\maketitle
\fancyhead[L]{}
\begin{abstract}
Low-rank adaptation (LoRA) enables parameter-efficient fine-tuning by learning low-rank weight updates. Increasing rank expands adaptation capacity, but does not guarantee that all learned directions improve downstream performance.  In a diagnostic experiment with rank-16 LoRA, we show that retaining only the top singular components achieves the best performance, whereas adding further components degrades it. Motivated by this, we propose Residual-Gradient Adaptation (ReGRA), a two-stage method for making more effective use of a fixed rank budget. After the first stage, ReGRA estimates task gradients on a calibration set and removes their projections onto the two-sided subspace induced by the first-stage update. It then aggregates the residual-gradient directions to initialize a second-stage adapter without changing the first-stage update. An orthogonality regularizer encourages the two updates to capture complementary directions. Both updates can be merged into the pretrained weights, incurring no additional inference overhead. Across diverse benchmark sets, ReGRA achieves stronger performance than alternatives with the same total rank and even matches LoRA adapters with twice the rank. Ablation studies and further analysis support the contribution of its core design choices. Code is available at \url{https://github.com/Fanyao-Li/ReGRA}.
\end{abstract}

\addtocontents{toc}{\protect\setcounter{tocdepth}{-1}}

\section{Introduction}

Large pretrained models have achieved strong performance across a wide range of downstream tasks, alongside scaling parameters and training corpus \citep{brown2020language,chen2021evaluating}. Efficient generation has also been explored through leap multi-token prediction \citep{liu2025lmtp}. In deployment, however, full-parameter fine-tuning for particular tasks or domains typically necessitates updating and storing full set of weights, motivating parameter-efficient fine-tuning (PEFT) \citep{houlsby2019parameter,he2022unified}. By optimizing only a small set of task-specific parameters, it significantly reduces the adaptation cost. Among PEFT methods, low-rank adaptation (LoRA) has become a widely used and effective choice, learning compact task-specific updates that can be merged into pretrained weights for inference~\citep{hu2022lora}.

The principle behind LoRA is to leverage low-rank trainable factors to approximate dense weight updates~\citep{aghajanyan2021intrinsic,hu2022lora}.
Specifically, for a pretrained weight $\mW_0$, LoRA parameterizes the task-specific update as $\Delta \mW=\mB\mA$, which satisfies $r=\text{rank}(\mB\mA)\ll d$, where $d$ denotes the dimension of $\mW$. $\mW_0$ remains frozen during the fine-tuning phase.
Therefore, it substantially reduces the number of trainable parameters, and the learned update can be merged into $W_0$ without additional inference overhead.
This balance of parameter efficiency and effective task specialization has kept LoRA widely used in resource-constrained settings.
Recent advances continue to refine its low-rank principle through adaptive rank allocation
~\citep{zhang2023adalora}, initialization informed by pretrained weights or task gradients
~\citep{meng2024pissa,wang2024loraga,zhang2025loraone}, optimization strategies tailored to its factorized parameterization
~\citep{hayou2024loraplus,wang2025lorapro} and adaptive adapter selection and placement~\citep{hayou2026plop,veprikov2026weightlora}.
Despite different methods to improve LoRA, they leave a fundamental question unanswered: \textit{does the learned update make effective use of its allocated rank?}

\begin{wrapfigure}{r}{0.52\linewidth}
    \centering
    \vspace{-4mm}
    \includegraphics[width=\linewidth,trim=40 15 10 30,clip]{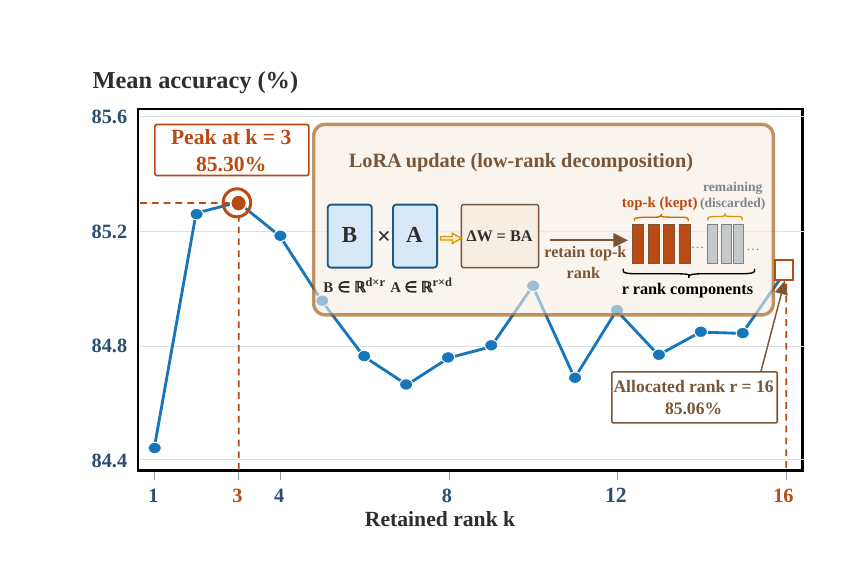}
    \vspace{-6mm}
    \caption{\textbf{Effect of retained rank on accuracy.} Model performance does not improve monotonically with increasing retained rank.}
    \vspace{-4mm}
    \label{fig:rank_analysis}
\end{wrapfigure}

To this end, we train a LoRA model with rank $r=16$ for diagnosis and decompose each learned update $\Delta\mW$ using singular value decomposition (SVD). Without retraining, we evaluate rank-$k$ approximations formed by retaining the top $k$ singular components. As shown in Figure~\ref{fig:rank_analysis},  we find that $k=3$ yields the best downstream performance, whereas retaining additional components degrades it. 
This suggests that the full learned update is \textit{less effective} than a lower-rank approximation of itself. A larger rank budget does not necessarily produce a more useful task-specific update. 

Motivated by this observation, we propose Residual-Gradient Adaptation (\textbf{ReGRA}), which helps the learned LoRA update make more effective use of its available rank. ReGRA first trains a standard LoRA adapter and then estimates task gradients with a small subset of the training data. To exhaust the allocated rank, we initialize a second-stage adapter with task gradients projected onto the two-sided complement of the first-stage update. With this first-stage estimation, we are able to leave the previously learned components unchanged, yet continually fulfil the update within the allocated rank. During second-stage training, we apply orthogonal regularization to maintain decorrelation between the two stages. After training, both updates can be merged into the base model, introducing no additional inference cost.
We evaluate ReGRA using standard LoRA evaluation following \citep{hu2023llmadapters} and expand it to natural language understanding, mathematical reasoning, and code generation. Under the same total-rank budget, ReGRA consistently achieves stronger overall performance across the evaluated benchmarks and even reaches the performance of the double-rank model. Ablation studies and further analysis support the effectiveness of residual-gradient initialization and update decorrelation.

Our contributions are summarized as follows:
\begin{itemize}
    \item We reveal that standard LoRA training does not always make effective use of its allocated rank. We approach this problem from a task-gradient perspective, targeting directions not captured by the initial update.
    \item We propose ReGRA to exhaust the allocated rank. ReGRA is a two-stage adaptation method that estimates task gradients after standard LoRA training, projects out the components captured by the first-stage update, and uses the residual gradients to initialize a second-stage adapter. An orthogonal regularization maintains the decorrelation between the two updates.
    \item We evaluate ReGRA on diverse benchmark sets, showing that it outperforms representative baselines and, in some settings, matches the performance of a model with twice the rank. Ablation studies support the efficacy and rationale of our design.
\end{itemize}

\section{Related Work}
\label{sec:related_work}

\noindent\textbf{Low-rank parameter-efficient fine-tuning.}
Low-rank adaptation (LoRA) represents task-specific weight updates using two trainable low-rank factors, reducing the number of trainable parameters while retaining mergeability at inference \citep{hu2022lora}. Subsequent variants improve its memory and parameter efficiency through quantization, factor freezing, or shared random projections \citep{dettmers2023qlora,liu2026freeact,zhang2023lorafa,kopiczko2024vera}. Other methods explore adaptive rank allocation and rank-dependent scaling \citep{zhang2023adalora,valipour2023dylora,kalajdzievski2023rslora}. LoRA+, DoRA, and LoRA-Pro instead improve the optimization or parameterization of low-rank updates \citep{hayou2024loraplus,liu2024dora,wang2025lorapro}. Recent works also explore adaptive adapter selection and placement to determine where low-rank adaptation is most effective ~\citep{hayou2026plop,veprikov2026weightlora}. Although these methods improve LoRA from different perspectives, they do not explicitly investigate redundancy among rank components within a learned update matrix.

\noindent\textbf{Spectral and gradient-informed initialization.}
Several methods initialize LoRA using information from pretrained weights, activations, or downstream data. PiSSA initializes adapters from the principal singular components of pretrained weights, whereas MiLoRA adapts their minor singular components \citep{meng2024pissa,wang2025milora}. CorDA constructs context-aware adaptation subspaces from weight and activation statistics, while EVA selects initialization directions according to activation variance \citep{yang2024corda,paischer2025eva}. Gradient-based methods use the task gradient at the start of fine-tuning to guide adapter initialization
\citep{wang2024loraga,zhang2025loraone,ponkshe2024lorasb}. These approaches determine the adaptation subspace before fine-tuning, whereas ReGRA estimates residual task gradients after a first-stage update has been learned.

\noindent\textbf{Sequential and subspace-aware adaptation.}
ReLoRA and Chain of LoRA (COLA) construct higher-rank effective updates by repeatedly training, merging, and reinitializing low-rank adapters \citep{lialin2024relora,xia2024cola}. LoRA-based adaptation has also been studied in continual learning.
CoLoR applies LoRA to domain-incremental learning\citep{wistuba2023color}, while O-LoRA
assigns sequential tasks to orthogonal low-rank subspaces and InfLoRA
constructs adaptation subspaces that reduce interference with prior tasks
\citep{wang2023olora,liang2024inflora}.
KeepLoRA further uses residual-gradient information to balance knowledge retention and plasticity across tasks \citep{luo2026keeplora}. CMCL uses dual-sided gradient projection to reduce forgetting in continual learning \citep{liu2025cmcl}. Unlike these methods, ReGRA considers two-stage adaptation to the same downstream objective and combines residual-gradient initialization with update decorrelation to encourage complementary stage-wise updates.

\begin{figure}[t]
    \centering
    \includegraphics[width=\linewidth, trim=20 0 20 0, clip]{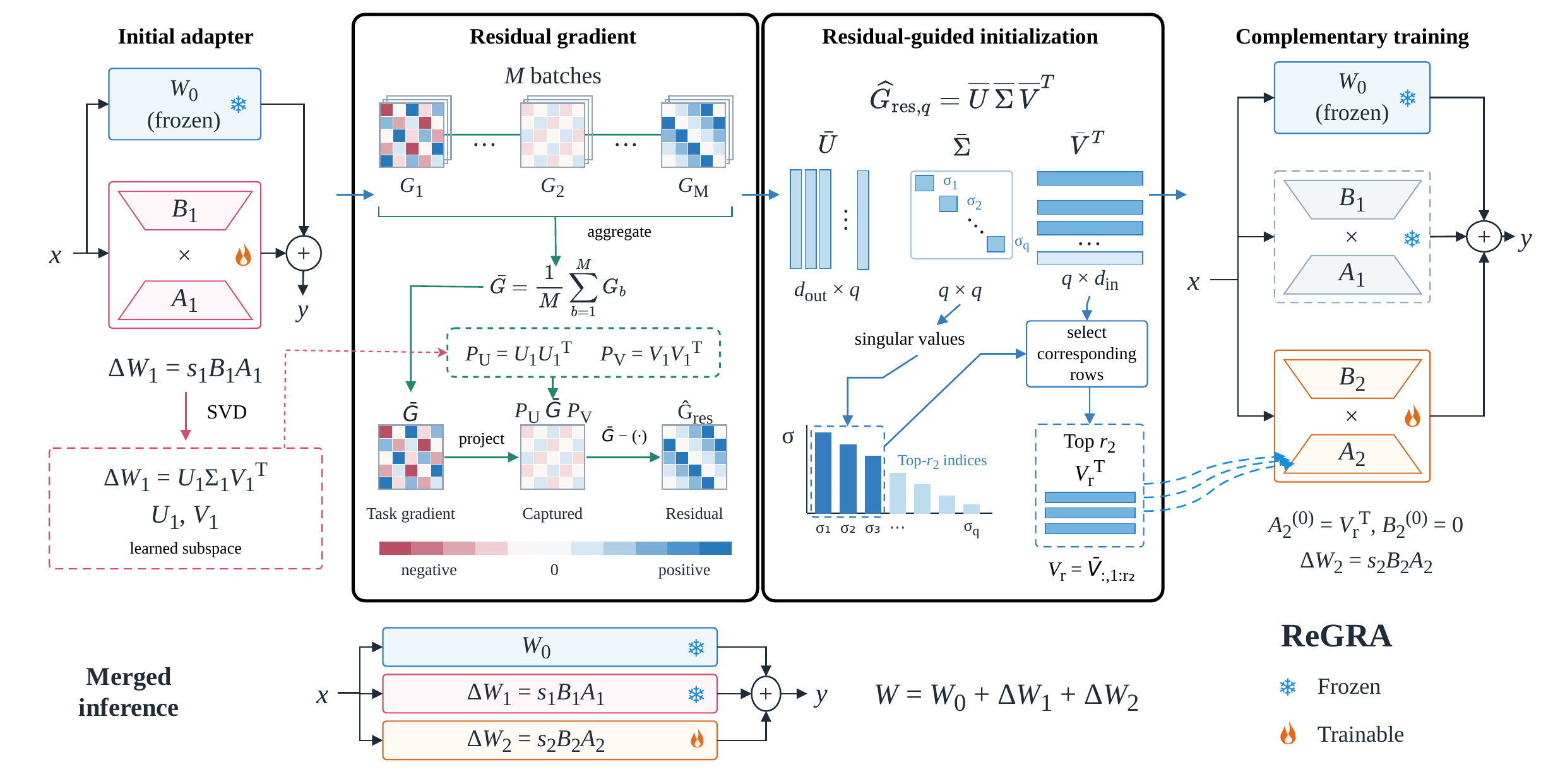}
    \vspace{-8mm}
    \caption{\textbf{Overall framework of Residual-Gradient Adaptation (ReGRA).} 
    ReGRA first trains Adapter~1 to learn an initial task-specific update.
It then projects the task gradient away from the subspace already captured by Adapter~1 and extracts the dominant residual directions via SVD.
These residual directions are used to initialize and guide Adapter~2 toward complementary task-relevant information.
The two adapters are finally combined with the frozen backbone to form the final model.
}
    \vspace{-4.5mm}
\label{fig:regra}
\end{figure}

\vspace{-2.5mm}
\section{Preliminaries}
\label{sec:preliminaries}

\leavevmode

\begin{wrapfigure}{r}{0.26\linewidth}
    \centering
    \vspace{-4mm}
    \includegraphics[width=\linewidth]{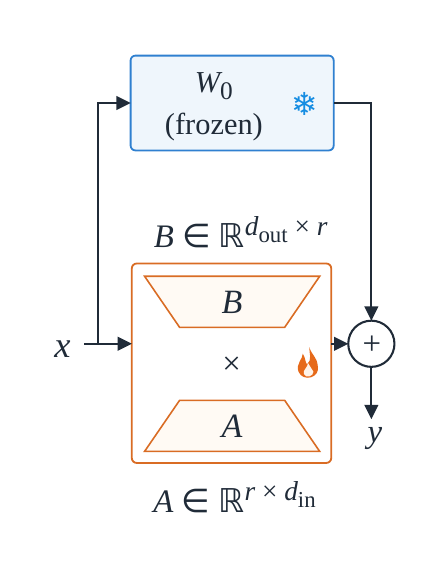}
    \vspace{-12mm}
    \caption{\textbf{Illustration of the standard LoRA.}}
    \label{fig:standard_lora}
    \vspace{-2mm}
\end{wrapfigure}

\noindent\textbf{Low-rank adaptation (LoRA).}
LoRA adapts a frozen pretrained weight
$\mW_0\in\mathbb{R}^{d_{\mathrm{out}}\times d_{\mathrm{in}}}$, as shown in Figure~\ref{fig:standard_lora}, by adding
a trainable low-rank update~\citep{hu2022lora}:
\begin{equation}
    \mW =
    \mW_0+\Delta\mW,
    \quad
    \Delta\mW = s\mB\mA.
    \label{eq:lora_update}
\end{equation}

Here,
$\mA\in\mathbb{R}^{r\times d_{\mathrm{in}}}$ and
$\mB\in\mathbb{R}^{d_{\mathrm{out}}\times r}$ are trainable factors,
$r$ is the adapter rank, and $s$ is the LoRA scaling coefficient.
The pretrained weight $\mW_0$ remains frozen during adaptation.
Several extensions modify how LoRA updates are trained or initialized.
QLoRA reduces memory use by quantizing the frozen backbone\citep{dettmers2023qlora}, while
AdaLoRA and DyLoRA explore flexible rank allocation
\citep{zhang2023adalora,valipour2023dylora}.
LoRA-GA uses the initial task gradient to guide adapter initialization
\citep{wang2024loraga}.
Our method focuses on learning a second-stage subspace that complements
the subspace learned in the first stage.

\noindent\textbf{Singular value decomposition (SVD).}
For a learned low-rank update $\Delta\mW$ with
$k=\operatorname{rank}(\Delta\mW)\leq r$, its compact singular value
decomposition is
\begin{equation}
    \Delta\mW
    =
    \mU\mSigma\mV^\top.
    \label{eq:compact_svd}
\end{equation}
The matrices
$\mU\in\mathbb{R}^{d_{\mathrm{out}}\times k}$ and
$\mV\in\mathbb{R}^{d_{\mathrm{in}}\times k}$ contain the left and right
singular vectors, respectively, while
$\mSigma=\operatorname{diag}(\sigma_1,\ldots,\sigma_k)$ contains the
positive singular values in descending order. The columns of $\mU$ span
the output-side directions represented by the update, whereas the columns
of $\mV$ span its input-side directions. These singular subspaces are used
to identify which components of a task gradient are already represented
by a learned low-rank update.

Spectral information has also been used to initialize LoRA adapters:
PiSSA selects the principal singular components of the pretrained weights\citep{meng2024pissa},
while MiLoRA selects the minor components
\citep{wang2025milora}.
Here, we apply SVD to the first-stage update $\Delta\mW$ to identify
the components of the task gradient already captured by the update.

\section{Residual-Gradient Adaptation}
\label{sec:method}
We briefly introduce ReGRA to provide an overall understanding. ReGRA decomposes adaptation into two sequential low-rank updates:
\begin{equation}
    \mW
    =
    \mW_0+\Delta\mW_1+\Delta\mW_2,
    \qquad
    \Delta\mW_i
    =
    s_i\mB_i\mA_i,
    \quad
    i\in\{1,2\}.
    \label{eq:regra_update}
\end{equation}
We describe ReGRA for a single adapted weight and omit the
layer index. The same procedure is applied independently to all target
weights. 
As demonstrated in Figure~\ref{fig:regra}, the first stage follows standard LoRA training, optimizing a rank-$r_1$
adapter with the objective 
$\mathcal{L}_{\mathrm{stage1}}=\mathcal{L}_{\mathrm{task}}
\left(
\mW_0+\Delta\mW_1;
\gD_{\mathrm{train}}
\right)$. ReGRA estimates task gradients and removes the component associated with the first stage, as characterized by the left and right singular subspaces of $\Delta\mW_1$. The residual gradient is used to initialize a rank-$r_2$ second stage adapter while preserving the function defined by $\mW_0+\Delta\mW_1$.
During second-stage training, we apply an update-level regularizer to promote orthogonality between the first- and second-stage updates.
After training, both updates can be merged into the
pretrained weight without introducing an additional inference-time branch.
The full procedure is listed in Appendix~\ref{app:algorithm-flow}.

\subsection{Residual-Guided Initialization}
\label{sec:residual_initialization}
To help the second stage learn complementary components, we first examine the first-stage update.
Applying the compact SVD to the fixed first-stage update gives
$\Delta\mW_1=\mU_1\mSigma_1\mV_1^\top$.
We define the projectors onto its output- and input-side subspaces as
$\mP_U=\mU_1\mU_1^\top$ and
$\mP_V=\mV_1\mV_1^\top$, respectively.
Let $\{\gC_b\}_{b=1}^{M}$ be calibration batches taken from the
training data. At the first-stage solution, we compute the task
gradient with respect to the adapted weight:
$\mG
=
\nabla_{\mW}\mathcal{L}_{\mathrm{task}}
(\mW_0+\Delta\mW_1;\gC_b).$
We remove its component in the two-sided subspace represented by
the first-stage update, as derived in Appendix~\ref{app:residual-orthogonality}:
\begin{equation}
\mG_{\mathrm{res}}
=
\mG-\mP_U\mG_b\mP_V.
\label{eq:residual_gradient}
\end{equation}
Let
$\widehat{\mG}_{\mathrm{res},q}
=
\overline{\mU}\overline{\mSigma}\overline{\mV}^{\top}$
denote the rank-$q$ truncated SVD approximation of
$\widehat{\mG}_{\mathrm{res}}$, , obtained by batchwise low-rank
aggregation as detailed in Appendix~\ref{app:gradient_aggregation}. 
The calibration procedure is described in Appendix~\ref{app:calibration-implementation}.
For a rank-$r_2$ second-stage adapter with $r_2\leq q$,
define
$\mV_r=\overline{\mV}_{:,1:r_2}$,
$\mU_r=\overline{\mU}_{:,1:r_2}$, and
$\mSigma_r=\overline{\mSigma}_{1:r_2,1:r_2}$.
ReGRA initializes the adapter as
\begin{equation}
\mA_2^{(0)}=\mV_r^\top,
\qquad
\mB_2^{(0)}=\mathbf{0}.
\label{eq:residual_initialization}
\end{equation}
This initialization specifies the adapter's input-side subspace
using the leading right singular vectors of the residual-gradient
estimate.
The following theorem characterizes its initial effect on
the model and its first task-gradient update.

\begin{proposition}[Residual-guided first-step update]
\label{thm:residual_initialization}
Consider $\Delta\mW_2=s_2\mB_2\mA_2$ initialized as in
Equation~\eqref{eq:residual_initialization}.
The initialization preserves the Stage-1 model function.
After one simultaneous gradient-descent step on the task loss,
the effective update lies in the input-side subspace spanned
by $\mV_r$. If the initial task gradient equals
$\widehat{\mG}_{\mathrm{res},q}$, this update is proportional
to its negative leading rank-$r_2$ component.
The proof is given in
Appendix~\ref{app:initialization_derivation}.
\end{proposition}

\begin{remarkbox}
\textbf{Remark.}
Since $\mB_2^{(0)}=\mathbf{0}$, we have
$\Delta\mW_2^{(0)}=\mathbf{0}$.
Let $\mG^{(0)}$ be the task gradient at
$\mW_0+\Delta\mW_1$. One simultaneous gradient-descent
step with learning rate $\eta$ gives
$\Delta\mW_2^{(1)}
=
-\eta\mG^{(0)}\mV_r\mV_r^\top.$
In the special case
$\mG^{(0)}=\widehat{\mG}_{\mathrm{res},q}$,
this reduces to
$\Delta\mW_2^{(1)}
=
-\eta\mU_r\mSigma_r\mV_r^\top.$
These expressions describe the first plain gradient-descent
step, not the entire Stage-2 trajectory. Both factors are
subsequently optimized, and the residual-gradient projection
is not applied to later task gradients.
\end{remarkbox}

\subsection{Stage-Wise Orthogonality Regularization}
\label{sec:update_decorrelation}

Residual-gradient initialization guides the second-stage adapter
toward directions associated with the remaining task gradient.
As training proceeds, the second-stage update can move away
from these initial directions. ReGRA therefore applies an
update-level regularizer to promote orthogonality between
the two stage-wise updates.
For each adapted weight, let
$\mD_1=\mB_1\mA_1$ and $\mD_2=\mB_2\mA_2$
denote the unscaled first- and second-stage updates.
We define the orthogonality penalty as
\begin{equation}
\Omega
=
\frac{
\langle\mD_1,\mD_2\rangle_F^2
}{
\lVert\mD_1\rVert_F^2
\lVert\mD_2\rVert_F^2+\varepsilon
},
\label{eq:layer_decorrelation}
\end{equation}
where $\varepsilon>0$ ensures numerical stability.
Squaring the inner product penalizes deviations from
orthogonality in either direction. The penalty promotes orthogonal updates.
For $L$ adapted weights, we average the penalties:
$\mathbb{R}_{\mathrm{decorr}}
=
\frac{1}{L}\sum_{\ell=1}^{L}\Omega_\ell.$
The inner products and norms can be computed from the
LoRA factors without forming dense update matrices.
The derivation is given in
Appendix~\ref{app:regularizer_derivation}.

During second-stage training, $\Delta\mW_1$ remains fixed,
while $\mA_2$ and $\mB_2$ are optimized using
\begin{equation}
\mathcal{L}_{\mathrm{stage2}}
=
\mathcal{L}_{\mathrm{task}}
\left(
\mW_0+\Delta\mW_1+\Delta\mW_2;
\gD_{\mathrm{train}}
\right)
+
\lambda_t\mathbb{R}_{\mathrm{decorr}}.
\label{eq:stage2_objective}
\end{equation}
We set $\lambda_t=0.1$ throughout the second stage
in all experiments.

\noindent\textbf{Computation and inference.}
Residual-gradient calibration introduces a one-time cost between the two
training stages. Each dense task gradient is formed only transiently,
projected, compressed to rank $q$, and then discarded, while cross-batch
aggregation and update decorrelation operate on low-rank factors. After
training, the two learned updates are merged into the pretrained weight as
$\mW_{\mathrm{merged}}
=
\mW_0+s_1\mB_1\mA_1+s_2\mB_2\mA_2$.
The Stage-2 procedure is detailed in Appendix~\ref{app:stage2-implementation}.
ReGRA therefore introduces neither an additional adapter branch nor extra
computation during inference, storage and computation bounds are given in Appendix~\ref{app:complexity}. 

\section{Experiments}
\label{sec:experiments}

To validate the efficacy and rationale of our design, we conduct experiments to address the following research questions (RQs):
\begin{itemize}[leftmargin=*]
    \item \textbf{RQ1:} How does ReGRA compare with baselines across diverse benchmark sets, and how does it compare with LoRA with larger rank capacity (\textit{i.e.}, $r=8$ \textit{v.s.} $r=16$)?
    \item \textbf{RQ2:} What are the separate and combined effects of residual-gradient initialization and update decorrelation?
    \item \textbf{RQ3:} Which singular components of a trained LoRA update affect performance when removed, and what do isolated components of ReGRA's second-stage update contribute?
    \item \textbf{RQ4:} How does performance change as rank-2 stages are progressively added?
\end{itemize}

\subsection{Experimental Setup}
\label{sec:experimental-setup}

\noindent\textbf{Tasks and datasets.}
We evaluate ReGRA on natural language understanding~\citep{wang2018glue},
mathematical reasoning~\citep{hu2023llmadapters,cobbe2021training},and code generation~\citep{chen2021evaluating,austin2021program}.
For natural language understanding, we fine-tune
RoBERTa-base and RoBERTa-large~\citep{liu2019roberta}
separately on the eight GLUE tasks: CoLA, SST-2, MRPC, STS-B, QQP, MNLI, QNLI, and RTE.
We evaluate all tasks on their labeled validation sets,
with the matched validation split used for MNLI.
For mathematical reasoning~\citep{hu2023llmadapters},
we fine-tune Llama-3-8B~\citep{grattafiori2024llama}
and Qwen3-8B-Base~\citep{yang2025qwen3}
on Math-7K training set
released with LLM-Adapters~\citep{hu2023llmadapters}.
We evaluate on AddSub, MultiArith, SingleEq,
GSM8K, AQuA, and SVAMP.
For code generation,
we fine-tune Llama-3-8B
on CodeFeedback-Filtered-Instruction~\citep{zheng2024opencodeinterpreter}
and evaluate on MBPP~\citep{austin2021program}
and its enhanced MBPP+ test suite using
EvalPlus~\citep{liu2023your}.

\noindent\textbf{Implementation details.}
Adapters are applied to the query, key, and value projections while the
pretrained backbones remain frozen, details are provided in
Appendix~\ref{app:model-training-setup}. Each ReGRA stage uses rank 8,
$\alpha=16$, and dropout 0.05; the GLUE classification head is trained in
the first stage and frozen in the second stage. Generative losses are
computed only on response tokens.  We use zero weight decay, and
$\lambda_{\mathrm{t}}=0.1$, whose contribution is warmed up over the first
100 second-stage steps. The training schedules for GLUE, mathematics, and
code generation comprise 20, 2, and 1 epochs, respectively, full settings are tabulated in Appendix~\ref{app:training-hyperparameters}.

\noindent\textbf{Baselines and evaluation.}
We compare ReGRA with LoRA~\citep{hu2022lora},
DoRA~\citep{liu2024dora},
PiSSA~\citep{meng2024pissa},
LoRA-One~\citep{zhang2025loraone},
and KeepLoRA~\citep{luo2026keeplora}.
LoRA is evaluated at ranks 8 and 16, while the other baselines use rank 8.
For ReGRA, $r=8$ refers to the rank of each stage.
GLUE uses the standard task-specific metrics:
Matthews correlation for CoLA, the mean of Pearson and Spearman correlations
for STS-B, the mean of accuracy and $F_1$ for MRPC and QQP, and accuracy
otherwise. Mathematical reasoning uses answer accuracy, and code generation
uses pass@1. All scores are reported as percentages.

\subsection{Main Results (RQ1)}
\label{sec:main-results}

\begin{table*}[t]
\caption{\textbf{Performance (\%) on eight GLUE natural language understanding tasks} using RoBERTa-base and RoBERTa-large. Avg. is the unweighted mean across tasks. We calculate the performance gain of ReGRA against the rank-8 LoRA on the same backbone. For ReGRA, $r=8$ denotes the rank of each stage. The best average for each backbone is marked in bold.}
\vspace{2mm}

\label{tab:glue-results}
\centering
\renewcommand{\arraystretch}{1.08}
\setlength{\tabcolsep}{2.2pt}

\definecolor{loraSixteenGray}{RGB}{125,125,125}
\definecolor{loraEightTint}{RGB}{240,240,240}
\definecolor{deltaGray}{RGB}{105,105,105}
\definecolor{regraTint}{RGB}{229, 245, 250}

\def\datasetHeader#1{%
    \makebox[2.8em][c]{\textbf{#1}}%
    \makebox[1.9em][l]{}%
}
\def\alignedScore#1{%
    \makebox[2.8em][r]{#1}%
    \makebox[1.9em][l]{}%
}

\def\scoreCell#1{\alignedScore{#1}}

\def\grayScoreCell#1{\cellcolor{loraEightTint}\alignedScore{#1}}

\def\loraSixteenScore#1{%
    \cellcolor{white}%
    \alignedScore{\textcolor{loraSixteenGray}{#1}}%
}

\def\regraCell#1#2{%
    \cellcolor{regraTint}%
    \makebox[2.8em][r]{#1}%
    \makebox[1.9em][l]{\ensuremath{{}_{\textcolor{deltaGray}{#2}}}}%
}

\resizebox{\textwidth}{!}{%
\begin{tabular}{@{}cl|cccccccc|c@{}}
\toprule
&
\textbf{Method}
& \datasetHeader{CoLA}
& \datasetHeader{MNLI}
& \datasetHeader{MRPC}
& \datasetHeader{QNLI}
& \datasetHeader{QQP}
& \datasetHeader{RTE}
& \datasetHeader{SST-2}
& \datasetHeader{STS-B}
& \datasetHeader{Avg.} \\
\midrule

\multirow{7}{*}{\rotatebox[origin=c]{90}{RoBERTa-base}}
& \textcolor{loraSixteenGray}{LoRA ($r=16$)}
& \loraSixteenScore{60.24}
& \loraSixteenScore{85.75}
& \loraSixteenScore{87.25}
& \loraSixteenScore{91.45}
& \loraSixteenScore{87.94}
& \loraSixteenScore{75.09}
& \loraSixteenScore{93.69}
& \loraSixteenScore{90.31}
& \loraSixteenScore{83.97} \\
\cmidrule(l){2-11}

& \cellcolor{loraEightTint}LoRA ($r=8$)
& \grayScoreCell{58.04}
& \grayScoreCell{85.11}
& \grayScoreCell{87.50}
& \grayScoreCell{90.99}
& \grayScoreCell{87.16}
& \grayScoreCell{72.92}
& \grayScoreCell{93.92}
& \grayScoreCell{90.07}
& \grayScoreCell{83.21} \\

& DoRA ($r=8$)
& \scoreCell{56.53}
& \scoreCell{86.20}
& \scoreCell{88.23}
& \scoreCell{91.21}
& \scoreCell{85.43}
& \scoreCell{74.36}
& \scoreCell{93.92}
& \scoreCell{90.05}
& \scoreCell{83.24} \\

& PiSSA ($r=8$)
& \scoreCell{56.01}
& \scoreCell{84.16}
& \scoreCell{89.12}
& \scoreCell{90.49}
& \scoreCell{87.43}
& \scoreCell{72.56}
& \scoreCell{93.23}
& \scoreCell{89.56}
& \scoreCell{82.82} \\

& LoRA-One ($r=8$)
& \scoreCell{58.39}
& \scoreCell{85.28}
& \scoreCell{89.12}
& \scoreCell{91.15}
& \scoreCell{85.51}
& \scoreCell{75.81}
& \scoreCell{94.38}
& \scoreCell{90.54}
& \scoreCell{83.77} \\

& KeepLoRA ($r=8$)
& \scoreCell{58.55}
& \scoreCell{84.76}
& \scoreCell{90.35}
& \scoreCell{90.93}
& \scoreCell{84.98}
& \scoreCell{67.87}
& \scoreCell{93.69}
& \scoreCell{90.42}
& \scoreCell{82.69} \\

\cmidrule(l){2-11}

& \cellcolor{regraTint}\textbf{ReGRA ($r=8$)}
& \regraCell{61.60}{+3.56}
& \regraCell{85.96}{+0.85}
& \regraCell{89.43}{+1.93}
& \regraCell{91.95}{+0.96}
& \regraCell{88.15}{+0.99}
& \regraCell{75.81}{+2.89}
& \regraCell{94.38}{+0.46}
& \regraCell{90.66}{+0.59}
& \regraCell{\textbf{84.74}}{+1.53} \\

\midrule

\multirow{7}{*}{\rotatebox[origin=c]{90}{RoBERTa-large}}
& \textcolor{loraSixteenGray}{LoRA ($r=16$)}
& \loraSixteenScore{64.55}
& \loraSixteenScore{90.59}
& \loraSixteenScore{90.33}
& \loraSixteenScore{94.58}
& \loraSixteenScore{87.57}
& \loraSixteenScore{85.92}
& \loraSixteenScore{96.22}
& \loraSixteenScore{92.20}
& \loraSixteenScore{87.75} \\
\cmidrule(l){2-11}

& \cellcolor{loraEightTint}LoRA ($r=8$)
& \grayScoreCell{66.02}
& \grayScoreCell{90.39}
& \grayScoreCell{89.70}
& \grayScoreCell{94.60}
& \grayScoreCell{88.93}
& \grayScoreCell{84.12}
& \grayScoreCell{96.22}
& \grayScoreCell{91.83}
& \grayScoreCell{87.73} \\

& DoRA ($r=8$)
& \scoreCell{65.53}
& \scoreCell{90.52}
& \scoreCell{91.04}
& \scoreCell{94.34}
& \scoreCell{87.53}
& \scoreCell{84.11}
& \scoreCell{96.10}
& \scoreCell{91.90}
& \scoreCell{87.63} \\

& PiSSA ($r=8$)
& \scoreCell{60.59}
& \scoreCell{89.89}
& \scoreCell{90.91}
& \scoreCell{94.08}
& \scoreCell{86.88}
& \scoreCell{83.75}
& \scoreCell{96.33}
& \scoreCell{91.72}
& \scoreCell{86.77} \\

& LoRA-One ($r=8$)
& \scoreCell{65.30}
& \scoreCell{90.35}
& \scoreCell{91.34}
& \scoreCell{94.63}
& \scoreCell{87.41}
& \scoreCell{85.19}
& \scoreCell{95.98}
& \scoreCell{92.30}
& \scoreCell{87.81} \\

& KeepLoRA ($r=8$)
& \scoreCell{60.72}
& \scoreCell{89.35}
& \scoreCell{91.71}
& \scoreCell{92.62}
& \scoreCell{86.04}
& \scoreCell{85.19}
& \scoreCell{94.61}
& \scoreCell{91.67}
& \scoreCell{86.49} \\

\cmidrule(l){2-11}

& \cellcolor{regraTint}\textbf{ReGRA ($r=8$)}
& \regraCell{68.82}{+2.80}
& \regraCell{90.45}{+0.06}
& \regraCell{91.76}{+2.06}
& \regraCell{94.60}{0.00}
& \regraCell{87.77}{-1.16}
& \regraCell{85.92}{+1.80}
& \regraCell{96.22}{0.00}
& \regraCell{91.85}{+0.02}
& \regraCell{\textbf{88.42}}{+0.69} \\

\bottomrule
\end{tabular}}
\vspace{-3mm}
\end{table*}

\begin{table*}[!t]
\caption{\textbf{Accuracy (\%) on six mathematical reasoning benchmarks} using
Llama-3-8B and Qwen3-8B-Base. Avg. is the unweighted mean across tasks.
Subscript values beside ReGRA scores show percentage-point differences from
rank-8 LoRA on the same backbone. For ReGRA, $r=8$ denotes the rank of each
stage. The best average for each backbone is marked in bold.}
\vspace{2mm}

\label{tab:math-results}
\centering
\renewcommand{\arraystretch}{1.08}
\setlength{\tabcolsep}{2.2pt}

\definecolor{loraSixteenGray}{RGB}{125,125,125}
\definecolor{loraEightTint}{RGB}{240,240,240}
\definecolor{deltaGray}{RGB}{105,105,105}
\definecolor{regraTint}{RGB}{229, 245, 250}

\def\datasetHeader#1{%
    \makebox[2.8em][c]{\textbf{#1}}%
    \makebox[2.2em][l]{}%
}
\def\alignedScore#1{%
    \makebox[2.8em][r]{#1}%
    \makebox[2.2em][l]{}%
}
\def\scoreCell#1{\alignedScore{#1}}
\def\grayScoreCell#1{\cellcolor{loraEightTint}\alignedScore{#1}}
\def\loraSixteenScore#1{%
    \cellcolor{white}%
    \alignedScore{\textcolor{loraSixteenGray}{#1}}%
}
\def\regraCell#1#2{%
    \cellcolor{regraTint}%
    \makebox[2.8em][r]{#1}%
    \makebox[2.2em][l]{\ensuremath{{}_{\textcolor{deltaGray}{#2}}}}%
}

\resizebox{0.96\textwidth}{!}{%
\begin{tabular}{@{}cl|cccccc|c@{}}
\toprule
& \textbf{Method}
& \datasetHeader{AddSub}
& \datasetHeader{MultiArith}
& \datasetHeader{SingleEq}
& \datasetHeader{SVAMP}
& \datasetHeader{GSM8K}
& \datasetHeader{AQuA}
& \datasetHeader{Avg.}
\\
\midrule
\multirow{7}{*}{\rotatebox[origin=c]{90}{Llama-3-8B}} & \textcolor{loraSixteenGray}{LoRA ($r=16$)} & \loraSixteenScore{79.74} & \loraSixteenScore{92.83} & \loraSixteenScore{87.59} & \loraSixteenScore{69.80} & \loraSixteenScore{60.50} & \loraSixteenScore{39.37} & \loraSixteenScore{71.64} \\

\cmidrule(l){2-9}
 & \cellcolor{loraEightTint}LoRA ($r=8$) & \grayScoreCell{76.20} & \grayScoreCell{92.66} & \grayScoreCell{87.99} & \grayScoreCell{69.30} & \grayScoreCell{61.41} & \grayScoreCell{37.40} & \grayScoreCell{70.83} \\

 & DoRA ($r=8$) & \scoreCell{77.72} & \scoreCell{94.16} & \scoreCell{86.41} & \scoreCell{72.10} & \scoreCell{60.34} & \scoreCell{39.76} & \scoreCell{71.75} \\

 & PiSSA ($r=8$) & \scoreCell{74.93} & \scoreCell{89.00} & \scoreCell{85.23} & \scoreCell{67.80} & \scoreCell{58.07} & \scoreCell{43.30} & \scoreCell{69.72} \\

 & LoRA-One ($r=8$) & \scoreCell{79.24} & \scoreCell{91.33} & \scoreCell{87.99} & \scoreCell{70.30} & \scoreCell{60.65} & \scoreCell{40.15} & \scoreCell{71.61} \\

 & KeepLoRA ($r=8$) & \scoreCell{80.50} & \scoreCell{87.33} & \scoreCell{88.18} & \scoreCell{70.50} & \scoreCell{60.80} & \scoreCell{37.79} & \scoreCell{70.85} \\

\cmidrule(l){2-9}
 & \cellcolor{regraTint}\textbf{ReGRA ($r=8$)} & \regraCell{79.24}{+3.04} & \regraCell{89.50}{-3.16} & \regraCell{88.19}{+0.20} & \regraCell{72.40}{+3.10} & \regraCell{62.55}{+1.14} & \regraCell{40.94}{+3.54} & \regraCell{\textbf{72.14}}{+1.31} \\

\midrule
\multirow{7}{*}{\rotatebox[origin=c]{90}{Qwen3-8B-Base}} & \textcolor{loraSixteenGray}{LoRA ($r=16$)} & \loraSixteenScore{91.65} & \loraSixteenScore{98.50} & \loraSixteenScore{95.08} & \loraSixteenScore{85.90} & \loraSixteenScore{77.41} & \loraSixteenScore{61.42} & \loraSixteenScore{84.99} \\

\cmidrule(l){2-9}
 & \cellcolor{loraEightTint}LoRA ($r=8$) & \grayScoreCell{91.39} & \grayScoreCell{98.50} & \grayScoreCell{94.09} & \grayScoreCell{86.00} & \grayScoreCell{75.28} & \grayScoreCell{59.84} & \grayScoreCell{84.18} \\

 & DoRA ($r=8$) & \scoreCell{91.65} & \scoreCell{98.83} & \scoreCell{94.29} & \scoreCell{84.80} & \scoreCell{75.97} & \scoreCell{62.99} & \scoreCell{84.76} \\

 & PiSSA ($r=8$) & \scoreCell{89.37} & \scoreCell{98.33} & \scoreCell{94.29} & \scoreCell{85.40} & \scoreCell{76.19} & \scoreCell{61.02} & \scoreCell{84.10} \\

 & LoRA-One ($r=8$) & \scoreCell{92.15} & \scoreCell{99.00} & \scoreCell{94.88} & \scoreCell{84.90} & \scoreCell{76.12} & \scoreCell{60.63} & \scoreCell{84.61} \\

 & KeepLoRA ($r=8$) & \scoreCell{86.84} & \scoreCell{97.50} & \scoreCell{94.09} & \scoreCell{81.90} & \scoreCell{77.48} & \scoreCell{62.99} & \scoreCell{83.47} \\

\cmidrule(l){2-9}
 & \cellcolor{regraTint}\textbf{ReGRA ($r=8$)} & \regraCell{92.91}{+1.52} & \regraCell{98.67}{+0.17} & \regraCell{95.28}{+1.19} & \regraCell{85.70}{-0.30} & \regraCell{75.51}{+0.23} & \regraCell{63.30}{+3.46} & \regraCell{\textbf{85.23}}{+1.05} \\

\bottomrule
\end{tabular}}
\vspace{-2mm}
\end{table*}

As shown in Table~\ref{tab:glue-results}, ReGRA
achieves the highest average score on both backbones. Increasing the
rank of standard LoRA from 8 to 16 also improves its average score,
confirming that additional low-rank capacity is useful. ReGRA obtains
a further improvement at the same maximum update rank by guiding how
that capacity is used. After learning the first-stage update, ReGRA
initializes the second adapter from task-gradient directions outside
the subspace associated with that update. Update-level decorrelation
then discourages the two stage-wise updates from aligning during
training. Standard rank-16 LoRA optimizes all rank components jointly
without this explicit guidance toward complementary updates. The
comparison supports the benefit of directing added capacity toward
optimization directions left insufficiently covered by the first
stage. Gains on CoLA, MRPC, and RTE further support this pattern.

\subsection{Results on Specialized Capabilities}
\label{sec:specialized-results}

\begin{figure*}[!t]
\centering

\begin{minipage}[t]{0.49\textwidth}
\vspace{-3mm}
\centering

\captionof{table}{
\textbf{Pass@1 (\%) on the MBPP and enhanced MBPP+ tests} using
Llama-3-8B. For ReGRA, $r=8$ denotes the rank of each stage.
The best result in each column is marked in bold.
}
\label{tab:code-results}
\vspace{4mm}

\renewcommand{\arraystretch}{1.08}
\setlength{\tabcolsep}{2.8pt}

\definecolor{loraSixteenGray}{RGB}{125,125,125}
\definecolor{loraEightTint}{RGB}{240,240,240}
\definecolor{deltaGray}{RGB}{105,105,105}
\definecolor{regraTint}{RGB}{229, 245, 250}

\def\datasetHeader#1{%
    \makebox[2.8em][c]{\textbf{#1}}%
    \makebox[2.0em][l]{}%
}
\def\alignedScore#1{%
    \makebox[2.8em][r]{#1}%
    \makebox[2.0em][l]{}%
}
\def\scoreCell#1{%
    \alignedScore{#1}%
}
\def\grayScoreCell#1{%
    \cellcolor{loraEightTint}%
    \alignedScore{#1}%
}
\def\loraSixteenScore#1{%
    \cellcolor{white}%
    \alignedScore{\textcolor{loraSixteenGray}{#1}}%
}
\def\regraCell#1#2{%
    \cellcolor{regraTint}%
    \makebox[2.8em][r]{#1}%
    \makebox[2.0em][l]{%
        \ensuremath{{}_{\textcolor{deltaGray}{#2}}}%
    }%
}

\resizebox{\linewidth}{!}{%
\begin{tabular}{@{}cl|cc@{}}
\toprule
\textbf{Backbone}
& \textbf{Method}
& \datasetHeader{MBPP}
& \datasetHeader{MBPP+}
\\
\midrule

\multirow{7}{*}{\rotatebox[origin=c]{90}{Llama-3-8B}}
& \textcolor{loraSixteenGray}{LoRA ($r=16$)}
& \loraSixteenScore{67.98}
& \loraSixteenScore{57.14}
\\

\cmidrule(l){2-4}

& \cellcolor{loraEightTint}LoRA ($r=8$)
& \grayScoreCell{65.87}
& \grayScoreCell{55.02}
\\

& DoRA ($r=8$)
& \scoreCell{66.40}
& \scoreCell{55.55}
\\

& PiSSA ($r=8$)
& \scoreCell{66.93}
& \scoreCell{54.49}
\\

& LoRA-One ($r=8$)
& \scoreCell{66.13}
& \scoreCell{54.76}
\\

& KeepLoRA ($r=8$)
& \scoreCell{67.46}
& \scoreCell{56.35}
\\

\cmidrule(l){2-4}

& \cellcolor{regraTint}\textbf{ReGRA ($r=8$)}
& \regraCell{\textbf{68.78}}{+2.91}
& \regraCell{\textbf{57.41}}{+2.39}
\\

\bottomrule
\end{tabular}
}

\end{minipage}%
\hfill%
\begin{minipage}[t]{0.47\textwidth}
\vspace{-1mm}
\centering

\includegraphics[
    width=0.95\linewidth,
    trim=0 28pt 0 12pt,
    clip
]{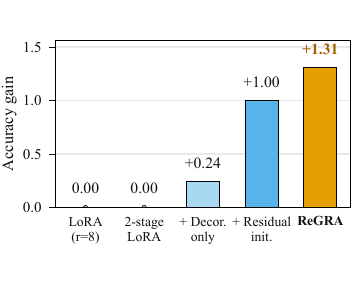}

\vspace{-10pt}
\captionof{figure}{
\textbf{Component ablation of ReGRA.}
Llama-3-8B gains in six-task average accuracy,
in percentage points, relative to LoRA ($r=8$).
Each configuration used one seed.
}
\label{fig:component-ablation}

\end{minipage}

\vspace{-4pt}
\end{figure*}

\noindent\textbf{Mathematical reasoning (RQ1).}
Table~\ref{tab:math-results} shows that ReGRA achieves the highest
six-task average on both Llama-3-8B and Qwen3-8B-Base. The task-level
pattern differs between the backbones. On Llama-3-8B, the gains over
rank-8 LoRA are concentrated on AddSub, SVAMP, and AQuA, while
MultiArith declines. On Qwen3-8B-Base, AddSub, SingleEq, and AQuA
benefit most, whereas ReGRA does not lead on every task. Its average
also exceeds rank-16 LoRA on both backbones, though the margin is
small on Qwen3-8B-Base. These results show that the overall
mathematical-reasoning benefit carries across two generative
backbones, with different task-level patterns.

\noindent\textbf{Code generation (RQ1).}
Table~\ref{tab:code-results} shows that ReGRA obtains the highest
pass@1 on both the base and enhanced MBPP+ tests. All methods score
lower on the enhanced tests, but ReGRA retains its lead under the
stricter evaluation. Its advantage over rank-16 LoRA is modest on
both test variants, so the code result supports a consistent
improvement within this benchmark rather than a large gain in code generation. 
The commonsense comparison is reported in
Appendix~\ref{app:commonsense-results}.

\subsection{Ablation Analysis (RQ2) }
\label{sec:ablation}

To disentangle the effects of the two key design components of ReGRA,
we conduct a controlled $2\times2$ ablation study on the mathematical
reasoning benchmarks. All variants share the same rank-8 first-stage adapter
and train an additional rank-8 Stage-2 adapter under identical optimization
settings. When residual-gradient initialization is removed, the second-stage
adapter adopts the standard LoRA initialization, with a randomly initialized
input-side factor and a zero-initialized output-side factor. When
update-level decorrelation is removed, the corresponding regularization term
is omitted from the Stage-2 objective. Performance is measured by the
unweighted average accuracy across the six mathematical reasoning datasets.

As shown in Figure~\ref{fig:component-ablation}, residual-gradient
initialization provides the dominant improvement, supporting the motivation
of directing second-stage adaptation toward gradient information insufficiently
captured by the first-stage. Update-level decorrelation provides a smaller additional
benefit by discouraging redundancy between the two updates. The full model
achieves the strongest overall performance, suggesting that the two components
play complementary roles: the former determines informative adaptation
directions, while the latter encourages the resulting update to remain distinct
from the first-stage.

\subsection{More Analysis}

\begin{figure}[t]
    \centering
    \includegraphics[
        width=0.97\linewidth,
        trim=0 10 0 15,
        clip
    ]{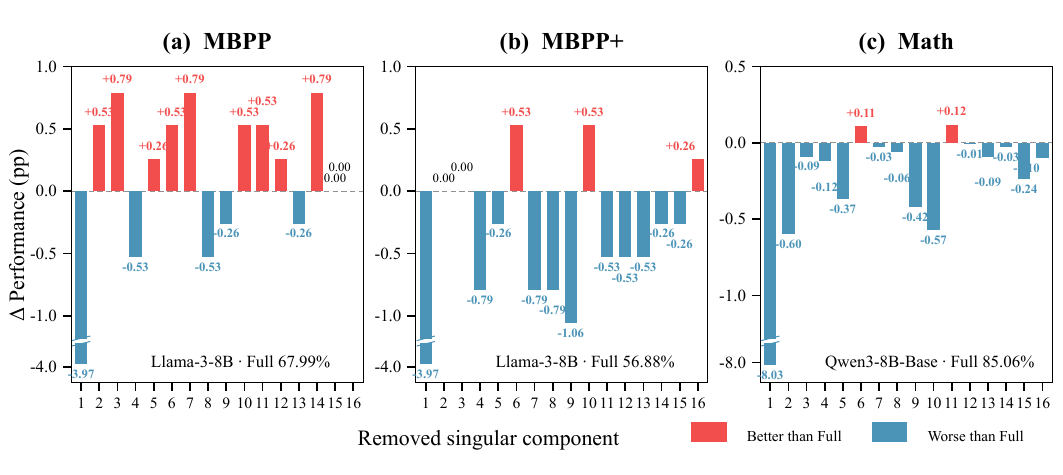}
    \vspace{-4mm}
    \caption{\textbf{Functional importance of LoRA singular components.}
For rank-16 LoRA, we apply SVD to each learned update $\Delta\mW$ and
remove one component at a time. Bars show performance changes relative
to the full update (dashed zero line) for Llama-3-8B on MBPP (a) and
MBPP+ (b), and Qwen3-8B-Base on Math (c). Removing the leading component
reduces performance by 3.97 percentage points on both code tasks and
8.03 percentage points on Math. Red and blue indicate gains and losses,
respectively; broken axes display the larger leading-component effect.}
    \label{fig:leave_one}
\end{figure}

\noindent\textbf{Unequal Functional Contributions of LoRA Singular Components (RQ3).}
We further examine how individual directions in a trained rank-16 LoRA
update affect downstream performance. We decompose each learned update
using SVD, remove one singular component at a time, and evaluate the
resulting model. As shown in Figure~\ref{fig:leave_one}, removing the
leading singular component produces the largest performance drop in
all three evaluations: 3.97 percentage points on both MBPP and MBPP+,
and 8.03 percentage points on Math. Removing most of the remaining
components has a much smaller effect, and removing some even improves
performance slightly.

These leave-one-component-out results measure the functional effect
of each removal, beyond the magnitude of its singular value. Together
with the top-$k$ analysis in Figure~\ref{fig:rank_analysis}, they
indicate that the learned LoRA directions do not contribute uniformly
to task performance. This observation motivates ReGRA to seek
complementary directions from the residual gradient after the
first-stage update has been learned.

\FloatBarrier

\begin{figure*}[t]
    \centering
    \includegraphics[
        width=0.8\textwidth,
        trim=0 14 0 14,
        clip
    ]{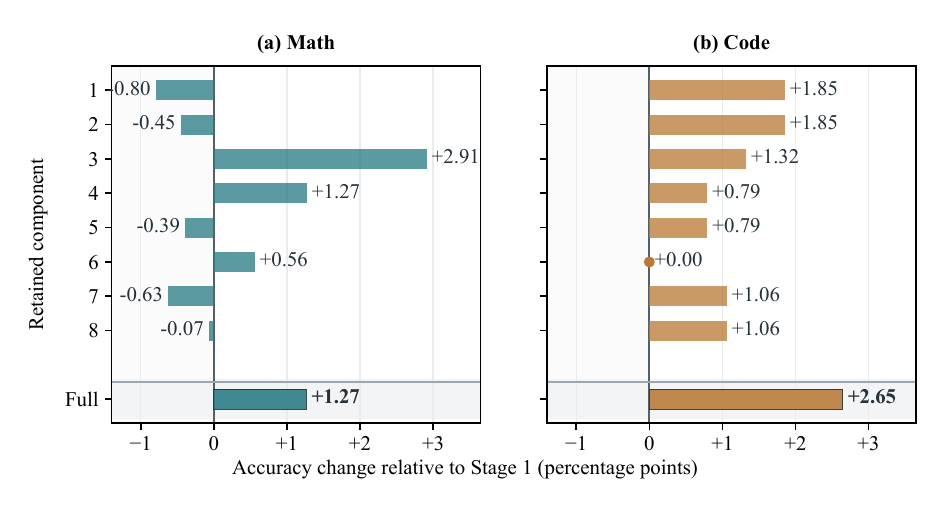}
    \vspace{-2mm}
    \caption{\textbf{Effects of isolated Stage-2 SVD components on Llama-3-8B.}
    Each $C_i$ retains only the $i$th rank-one component of every Stage-2
    update matrix on a fixed Stage-1 model; components are evaluated
    independently, not cumulatively. \emph{Full} retains all eight components.
    Bars show percentage-point changes from Stage 1-only in (a) mean accuracy
    across six math benchmarks and (b) MBPP+ pass\@1 on 378 problems. Positive
    and negative bars indicate gains and losses, respectively. Results are
    from one deterministic evaluation per condition.}
    \vspace{-14pt}
    \label{fig:stage2-component-performance}
\end{figure*}

\noindent\textbf{Functional diversity of Stage-2 components (RQ3).}
To examine the functional contribution of the learned Stage-2 update,
we decompose each trained Stage-2 update matrix by SVD and evaluate eight
independent variants. Variant $C_i$ retains only the $i$th rank-one component
in each adapted matrix while keeping Stage 1 fixed; the variants are not
cumulative. As shown in Figure~\ref{fig:stage2-component-performance},
their effects differ across task families. On the six mathematical reasoning
benchmarks, only $C_3$, $C_4$, and $C_6$ improve mean accuracy over
Stage 1-only, by 2.91, 1.27, and 0.56 percentage points, respectively;
the other five components produce small decreases. The full Stage-2 update
gains 1.27 points, below $C_3$ alone. On MBPP+, seven components improve
pass@1 by 0.79--1.85 points, while $C_6$ leaves it unchanged; the full
update gains 2.65 points. These results show heterogeneous and task-dependent
component effects. The full update's benchmark performance is not a sum of
its isolated component effects, although these independent evaluations do
not identify each component's marginal contribution within the full update, example-level error analyses are reported in
Appendix~\ref{app:stage2-error-analysis}.

\begin{wrapfigure}{r}{0.57\textwidth}
    \centering
    \vspace{-15pt}
    \includegraphics[width=\linewidth, trim=110 0 110 0, clip]{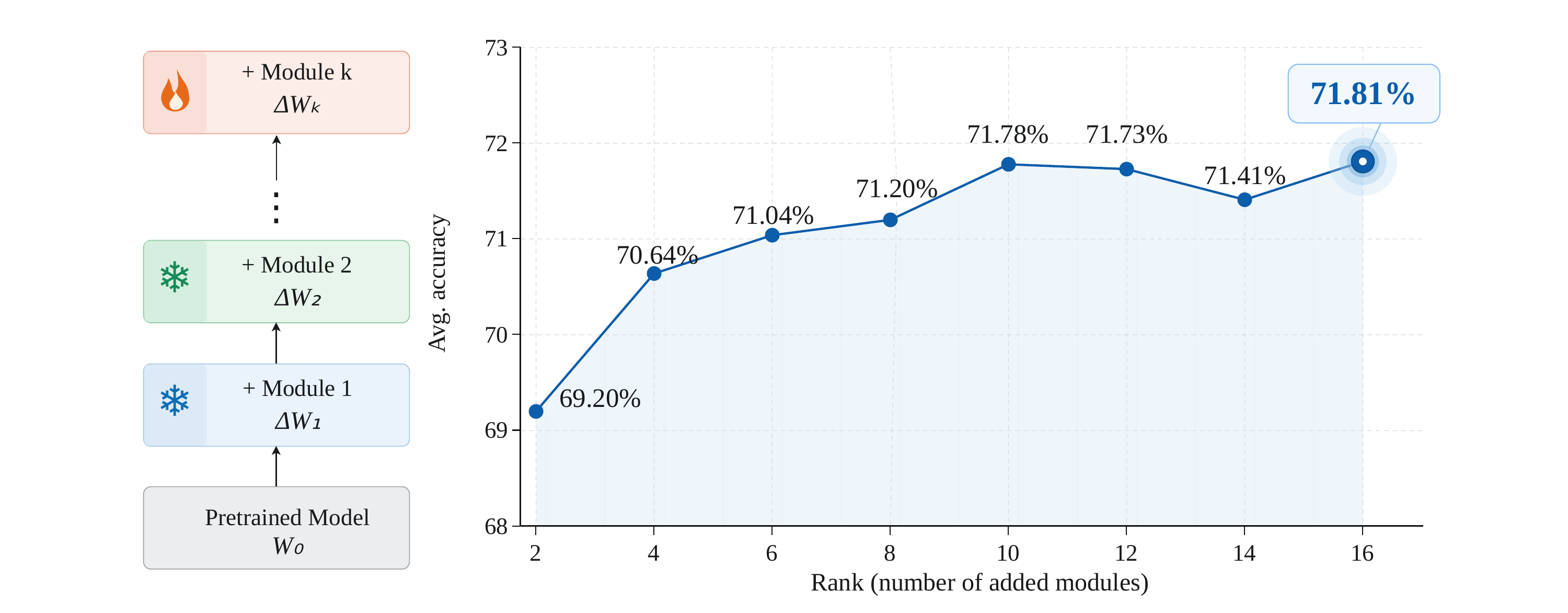}
    \vspace{-19pt}
    \caption{
    \textbf{Progressive rank expansion} on LLaMA-3-8B. At each stage, previously learned adapters are frozen and a new rank-2 adapter is trained.
    }
    \label{fig:rank_expansion}
    \vspace{-10pt}
\end{wrapfigure}

\noindent\textbf{Progressive Rank Expansion (RQ4).}
We further evaluate our method on LLaMA-3-8B under a multi-stage training setting. Starting from the base model, we iteratively train a new rank-2 adapter at each stage while freezing all previously learned adapters. Therefore, the effective adaptation rank increases progressively as more adapters are accumulated. As shown in Fig.~\ref{fig:rank_expansion}, the overall performance consistently improves with the increasing cumulative rank, rising from 69.20\% at rank 2 to 71.81\% at rank 16. Although minor fluctuations are observed at intermediate ranks, the overall trend demonstrates that newly introduced adapters continue to provide complementary adaptation capacity rather than merely duplicating previously learned directions. These results support the effectiveness of our additive multi-stage training strategy for progressively expanding the model's adaptation capacity.

\section{Conclusion}
\label{sec:conclusion}
In this work, we introduced ReGRA, a two-stage adaptation method that makes more effective use of a fixed total-rank budget. Motivated by the observation that additional LoRA singular components do not always improve performance, ReGRA uses residual task gradients to initialize a second adapter while preserving the first-stage solution. An orthogonality regularizer helps maintain separation between the two updates, which can be merged without additional inference overhead. Experiments on natural language understanding, mathematical reasoning, and code generation demonstrate improvements over rank-matched LoRA, while ablations support both design choices. These findings suggest that effective low-rank adaptation depends not only on increasing rank, but also on learning useful additional components relative to an existing update. Future work may investigate more principled ways to identify and allocate useful residual directions under a fixed rank budget.

\bibliography{iclr2027_conference}
\bibliographystyle{iclr2027_conference}

\addtocontents{toc}{\protect
\setcounter{tocdepth}{2}}

\newpage
\appendix
\onecolumn
\begin{center}
\LARGE \textbf{Appendix}
\end{center}
\tableofcontents
\let
\addcontentsline\OriginalAddContentsLine
\vspace{10mm}

\section{Theoretical Analysis}
\label{app:theoretical-analysis}

In this section, we analyze the residual-gradient projection
(Appendix~\ref{app:residual-orthogonality}),
its low-rank aggregation
(Appendix~\ref{app:gradient_aggregation}),
the initial Stage-2 update
(Appendix~\ref{app:initialization_derivation}),
and the factorized decorrelation penalty
(Appendix~\ref{app:regularizer_derivation}).

\subsection{Residual-Gradient Projection and Orthogonality}
\label{app:residual-orthogonality}

Consider an adapted weight
$\mW_0\in\mathbb{R}^{d_{\mathrm{out}}\times d_{\mathrm{in}}}$.
For the fixed first-stage update, define
\begin{equation}
\begin{aligned}
\Delta\mW_1
    &=s_1\mB_1\mA_1
      =\mU_1\mSigma_1\mV_1^\top,\\
\mP_U&=\mU_1\mU_1^\top,
\qquad
\mP_V=\mV_1\mV_1^\top .
\end{aligned}
\label{eq:app_projectors}
\end{equation}
For a calibration batch $\gC_b$, the gradient and its residual are
\begin{equation}
\begin{aligned}
\mG_b
    &=
    \nabla_{\mW}\mathcal{L}_{\mathrm{task}}
    (\mW_0+\Delta\mW_1;\gC_b),\\
\mG_{\mathrm{res},b}
    &=\mG_b-\mP_U\mG_b\mP_V\\
    &=\mG_b-
      \mU_1\mU_1^\top\mG_b
      \mV_1\mV_1^\top .
\end{aligned}
\label{eq:app_batch_residual}
\end{equation}
The residual is orthogonal to every matrix in the two-sided
subspace represented by the first-stage update. Indeed, for any
compatible $\mZ$,
\begin{equation}
\begin{aligned}
\left\langle
\mG_{\mathrm{res},b},
\mU_1\mZ\mV_1^\top
\right\rangle_F
&=
\left\langle
\mU_1^\top\mG_{\mathrm{res},b}\mV_1,
\mZ
\right\rangle_F\\
&=
\left\langle
\mU_1^\top
\bigl(\mG_b-\mP_U\mG_b\mP_V\bigr)
\mV_1,\mZ
\right\rangle_F\\
&=
\left\langle
\mU_1^\top\mG_b\mV_1
-\mU_1^\top\mG_b\mV_1,
\mZ
\right\rangle_F\\
&=0.
\end{aligned}
\label{eq:app_residual_orthogonality}
\end{equation}
Thus, the projection removes components jointly aligned with the
left and right first-stage subspaces. Unlike
$(\mI-\mP_U)\mG_b(\mI-\mP_V)$, it retains components aligned
with only one of these subspaces.

\subsection{Low-Rank Calibration and Gradient Aggregation}
\label{app:gradient_aggregation}

For each calibration batch, we truncate the residual gradient
immediately after projection. Let $\mathcal{T}_q(\cdot)$ denote
the rank-at-most-$q$ matrix returned by the low-rank SVD routine.
Then
\begin{equation}
\begin{aligned}
\mG_{\mathrm{res},b}
&=\mG_b-\mP_U\mG_b\mP_V,\\
\widetilde{\mG}_{\mathrm{res},b}
&:=\mathcal{T}_q
  \bigl(\mG_{\mathrm{res},b}\bigr)\\
&=\mU_b^{(g)}\mSigma_b^{(g)}
  \bigl(\mV_b^{(g)}\bigr)^\top,\\
\widehat{\mG}_{\mathrm{res}}
&:=\frac{1}{M}\sum_{b=1}^{M}
  \widetilde{\mG}_{\mathrm{res},b}\\
&=\frac{1}{M}\sum_{b=1}^{M}
  \mU_b^{(g)}\mSigma_b^{(g)}
  \bigl(\mV_b^{(g)}\bigr)^\top .
\end{aligned}
\label{eq:app_residual_mean}
\end{equation}
The superscript $(g)$ distinguishes the batch-gradient singular
vectors from those of the first-stage update.
The mean can be aggregated without constructing a dense matrix.
Define the concatenated factors
\begin{equation}
\begin{aligned}
\mathbf L
&=
\left[
\frac{\mU_1^{(g)}\mSigma_1^{(g)}}{M},
\ldots,
\frac{\mU_M^{(g)}\mSigma_M^{(g)}}{M}
\right],\\
\mathbf R
&=
\left[
\mV_1^{(g)},\ldots,\mV_M^{(g)}
\right].
\end{aligned}
\label{eq:app_concatenated_factors}
\end{equation}
Using thin QR factorizations
$\mathbf L=\mathbf Q_L\mathbf R_L$ and
$\mathbf R=\mathbf Q_R\mathbf R_R$, we obtain
\begin{equation}
\begin{aligned}
\widehat{\mG}_{\mathrm{res}}
&=\mathbf L\mathbf R^\top\\
&=(\mathbf Q_L\mathbf R_L)
  (\mathbf Q_R\mathbf R_R)^\top\\
&=\mathbf Q_L
  (\mathbf R_L\mathbf R_R^\top)
  \mathbf Q_R^\top .
\end{aligned}
\label{eq:app_factorized_mean}
\end{equation}
Thus, the SVD can be performed on the smaller core matrix
$\mathbf K=\mathbf R_L\mathbf R_R^\top$.
Let $\mathbf K_q$ be its rank-$q$ truncated SVD. The
corresponding truncated aggregate and Stage-2 initialization are
\begin{equation}
\begin{aligned}
\widehat{\mG}_{\mathrm{res},q}
&:=\mathbf Q_L\mathbf K_q\mathbf Q_R^\top\\
&=\overline{\mU}\overline{\mSigma}
  \overline{\mV}^{\top},\\
\mA_2^{(0)}
&=\bigl(\overline{\mV}_{:,1:r_2}\bigr)^\top,
\qquad
\mB_2^{(0)}=\mathbf{0},
\end{aligned}
\label{eq:app_aggregated_initialization}
\end{equation}
where $r_2\leq q$. This procedure obtains the initialization
directions without explicitly forming the dense aggregate
$\widehat{\mG}_{\mathrm{res}}$.
The orthogonality in
Equation~\eqref{eq:app_residual_orthogonality} applies to each
residual gradient before truncation.

\subsection{Proof of the Residual-Guided First-Step Update}
\label{app:initialization_derivation}

We prove Theorem~\ref{thm:residual_initialization} for one
adapted weight. Write
\begin{equation}
\begin{aligned}
\Delta\mW_2^{(0)}
&=s_2\mB_2^{(0)}\mA_2^{(0)}\\
&=s_2\mathbf{0}\mV_r^\top\\
&=\mathbf{0},
\end{aligned}
\label{eq:app_zero_initial_update}
\end{equation}
so the initial Stage-2 model represents exactly
$\mW_0+\Delta\mW_1$.

Let $\mG^{(0)}$ be the task-loss gradient with respect to the
adapted weight at this initial model. The chain rule gives
\begin{equation}
\begin{aligned}
\nabla_{\mB_2}\mathcal{L}_{\mathrm{task}}
    &=s_2\mG\mA_2^\top,\\
\nabla_{\mA_2}\mathcal{L}_{\mathrm{task}}
    &=s_2\mB_2^\top\mG .
\end{aligned}
\label{eq:app_factor_gradients}
\end{equation}
Because $\mB_2^{(0)}=\mathbf{0}$, one simultaneous
gradient-descent step with learning rate $\eta$ gives
\begin{equation}
\begin{aligned}
\mA_2^{(1)}
&=\mA_2^{(0)}
-\eta s_2\bigl(\mB_2^{(0)}\bigr)^\top\mG^{(0)}
=\mV_r^\top,\\
\mB_2^{(1)}
&=\mB_2^{(0)}
-\eta s_2\mG^{(0)}
 \bigl(\mA_2^{(0)}\bigr)^\top\\
&=-\eta s_2\mG^{(0)}\mV_r .
\end{aligned}
\label{eq:app_first_factor_step}
\end{equation}
Therefore,
\begin{equation}
\begin{aligned}
\Delta\mW_2^{(1)}
&=s_2\mB_2^{(1)}\mA_2^{(1)}\\
&=s_2
  \bigl(-\eta s_2\mG^{(0)}\mV_r\bigr)
  \mV_r^\top\\
&=-\eta s_2^2
  \mG^{(0)}\mV_r\mV_r^\top .
\end{aligned}
\label{eq:app_first_weight_step}
\end{equation}
In the special case
$\mG^{(0)}=\widehat{\mG}_{\mathrm{res},q}$,
the SVD definition of $\mV_r$ further yields
\begin{equation}
\begin{aligned}
\Delta\mW_2^{(1)}
&=-\eta s_2^2
  \widehat{\mG}_{\mathrm{res},q}
  \mV_r\mV_r^\top\\
&=-\eta s_2^2
  \overline{\mU}\overline{\mSigma}
  \overline{\mV}^{\top}
  \mV_r\mV_r^\top\\
&=-\eta s_2^2
  \mU_r\mSigma_r\mV_r^\top .
\end{aligned}
\label{eq:app_aligned_first_step}
\end{equation}
This result describes one plain gradient-descent step. It does
not assume that later mini-batch gradients equal the calibration
estimate or that subsequent updates remain in the initialized
subspace.

\subsection{Update Orthogonality Regularization}
\label{app:regularizer_derivation}

Let $\mD_1=\mB_1\mA_1$ and
$\mD_2=\mB_2\mA_2$ be the unscaled stage-wise updates.
Their Frobenius inner product satisfies
\begin{equation}
\begin{aligned}
\langle\mD_1,\mD_2\rangle_F
&=\operatorname{tr}(\mD_1^\top\mD_2)\\
&=\operatorname{tr}
  (\mA_1^\top\mB_1^\top\mB_2\mA_2)\\
&=\operatorname{tr}
  \bigl[(\mB_1^\top\mB_2)^\top
        (\mA_1\mA_2^\top)\bigr]\\
&=
\left\langle
\mB_1^\top\mB_2,\,
\mA_1\mA_2^\top
\right\rangle_F .
\end{aligned}
\label{eq:app_factorized_inner_product}
\end{equation}
Similarly,
\begin{equation}
\begin{aligned}
\|\mD_1\|_F^2
&=\operatorname{tr}(\mD_1^\top\mD_1)
 =\left\langle
   \mB_1^\top\mB_1,\,
   \mA_1\mA_1^\top
  \right\rangle_F,\\
\|\mD_2\|_F^2
&=\operatorname{tr}(\mD_2^\top\mD_2)
 =\left\langle
   \mB_2^\top\mB_2,\,
   \mA_2\mA_2^\top
  \right\rangle_F .
\end{aligned}
\label{eq:app_factorized_norms}
\end{equation}
Substitution into Equation~\eqref{eq:layer_decorrelation}
gives a factor-only expression:
\begin{equation}
\Omega=
\frac{
\left\langle
\mB_1^\top\mB_2,\,
\mA_1\mA_2^\top
\right\rangle_F^{\,2}
}{
\left\langle
\mB_1^\top\mB_1,\,
\mA_1\mA_1^\top
\right\rangle_F
\left\langle
\mB_2^\top\mB_2,\,
\mA_2\mA_2^\top
\right\rangle_F
+\varepsilon
}.
\label{eq:app_factorized_penalty}
\end{equation}
Thus, the penalty can be evaluated without explicitly forming
either dense update matrix. It helps maintain orthogonality between $\mD_1$ and $\mD_2$
during Stage-2 training; it does not require
their left or right singular subspaces to be separately
orthogonal.

\section{Supplementary Implementation Details}
\label{app:implementation}

This section specifies the model setup, residual-gradient
calibration, training hyperparameters, algorithm flow, and
computational cost of ReGRA.

\subsection{Model and Training Setup}
\label{app:model-training-setup}

We apply LoRA adapters to the query, key, and value projections.
The pretrained backbone remains fixed during both training stages.
Each ReGRA stage uses rank $8$, LoRA scaling parameter $\alpha=16$,
and dropout $0.05$. Stage 1 trains a standard LoRA adapter on the
task training data. Its learned update is then fixed during
calibration and Stage-2 training.

For GLUE, the task-specific classification head is trained in
Stage 1 and retained without further updates in Stage 2. For
generative tasks, the language-modeling loss is computed only
on response tokens. Experiments use seed $42$, bfloat16 precision,
and zero weight decay. The training runs were conducted on an
NVIDIA GeForce RTX 3090.

\subsection{Residual-Gradient Calibration}
\label{app:calibration-implementation}

Calibration is performed once after Stage 1 and before Stage 2.
We load the pretrained backbone together with the fixed Stage-1
adapter and evaluate task gradients on batches drawn from the
training data. The selected calibration records are processed
without data-loader shuffling. For each adapted weight, a
gradient with respect to the underlying weight is computed
without applying an optimizer update.

Each gradient is projected using the first-stage left and right
subspaces, as described in
Appendix~\ref{app:residual-orthogonality}. The residual is
immediately compressed to rank at most $q$, and the dense
gradient is discarded. Batchwise low-rank factors are aggregated
using the procedure in
Appendix~\ref{app:gradient_aggregation}. The leading right
singular vectors of the aggregate initialize $\mA_2$, while
$\mB_2$ is initialized to zero. Consequently, the initialized
Stage-2 adapter does not change the function represented by
the Stage-1 model.

For commonsense reasoning, the calibration subset contains
$64$ examples, with eight selected from each of the eight
tasks. This task balancing applies only to calibration; it
does not change the order of subsequent Stage-2 training.

\subsection{Hyperparameter Settings}
\label{app:training-hyperparameters}

Both stages use fused AdamW with a linear learning-rate
schedule and 100 optimizer-step learning-rate warmup. The
effective training batch sizes, maximum sequence lengths,
epoch counts, and peak learning rates are listed in
Table~\ref{tab:app-training-settings}. Both stages use the
same epoch count within a given task.

\begin{table}[t]
\centering
\caption{Training settings for ReGRA. The Stage-2 learning
rate for RoBERTa-large RTE is $10^{-4}$ rather than
$5\times10^{-5}$.}
\label{tab:app-training-settings}
\resizebox{\linewidth}{!}{%
\begin{tabular}{lccccc}
\toprule
Setting
& Epochs
& Stage-1 LR
& Stage-2 LR
& Batch size
& Max. length \\
\midrule
RoBERTa-base, GLUE
& Task-specific
& $2\times10^{-4}$
& $2\times10^{-4}$
& 32 & 128 \\
RoBERTa-large, GLUE
& Task-specific
& $2\times10^{-4}$
& $5\times10^{-5}$\textsuperscript{*}
& 32 & 128 \\
Llama-3-8B, math
& 2
& $10^{-4}$
& $10^{-4}$
& 2 & 256 \\
Qwen3-8B-Base, math
& 2
& $10^{-4}$
& $10^{-5}$
& 2 & 256 \\
Llama-3-8B, code
& 1
& $2\times10^{-5}$
& $5\times10^{-6}$
& 4 & 1024 \\
Llama-3-8B, commonsense
& 1
& $10^{-4}$
& $2\times10^{-5}$
& 4 & 256 \\
\bottomrule
\end{tabular}%
}

\vspace{2pt}
\raggedright
\textsuperscript{*}Except RTE, which uses $10^{-4}$.
\end{table}

For both RoBERTa backbones, CoLA, MNLI, MRPC, QNLI, QQP,
RTE, SST-2, and STS-B are trained for $10$, $3$, $20$, $3$,
$3$, $20$, $5$, and $10$ epochs, respectively.

Table~\ref{tab:app-calibration-settings} gives the calibration
settings. Here, $M$ is the number of calibration batches,
$b$ is their batch size, and $q$ is the gradient compression
rank. The low-rank SVD routine uses four iterations in all
listed configurations.

\begin{table}[t]
\centering
\caption{Residual-gradient calibration settings for ReGRA.}
\label{tab:app-calibration-settings}
\begin{tabular}{lccc}
\toprule
Setting & $M$ & $b$ & $q$ \\
\midrule
GLUE, both backbones       & 8  & 32 & 16 \\
Llama-3-8B, math           & 1  & 1  & 8  \\
Qwen3-8B-Base, math        & 4  & 2  & 8  \\
Llama-3-8B, code           & 16 & 1  & 16 \\
Llama-3-8B, commonsense    & 16 & 4  & 32 \\
\bottomrule
\end{tabular}
\end{table}

\subsection{Stage-2 Optimization and Merging}
\label{app:stage2-implementation}

During Stage 2, the backbone and Stage-1 adapter remain fixed.
Only the Stage-2 LoRA factors are optimized. Training uses
ordinary task gradients; the residual-gradient projection is
used for initialization rather than applied to every training
step. The task objective is augmented with the update-level
orthogonality regularizer in
Equation~\eqref{eq:layer_decorrelation}.

We linearly warm up the regularization coefficient over the
first 100 Stage-2 optimization steps:
\begin{equation}
\lambda_t
=
0.1\min\left\{1,\frac{t}{100}\right\},
\label{eq:app_regularizer_schedule}
\end{equation}
where $t$ denotes the Stage-2 optimization step. This
regularizer warmup is separate from the learning-rate warmup.

After training, the effective updates from both stages are
merged into each adapted pretrained weight:
\begin{equation}
\mW_{\mathrm{merged}}
=
\mW_0+\Delta\mW_1+\Delta\mW_2.
\label{eq:app_merged_weight}
\end{equation}
The merged model requires no additional adapter branch or
adapter-specific computation during inference.

\subsection{Algorithm Flow}
\label{app:algorithm-flow}

Algorithm~\ref{alg:regra} summarizes the complete procedure.
Stage 1 supplies a fixed task-specific update. Calibration
constructs a low-rank estimate of the residual gradient at
that solution. Stage 2 is initialized from this estimate,
optimized with update-orthogonality regularization, and
finally merged with Stage 1.

\begin{algorithm}[t]
\caption{ReGRA: two-stage residual-gradient adaptation}
\label{alg:regra}
\begin{algorithmic}[1]
\Require Pretrained weights
$\{\mW_{0,\ell}\}_{\ell=1}^{L}$;
training data $\gD_{\mathrm{train}}$;
calibration batches $\{\gC_b\}_{b=1}^{M}$;
stage ranks $r_1,r_2$; compression rank $q\geq r_2$
\Ensure Adapted weights
$\{\mW_{\ell}^{*}\}_{\ell=1}^{L}$

\State Train a rank-$r_1$ Stage-1 LoRA adapter on
$\gD_{\mathrm{train}}$
\State Fix its updates
$\{\Delta\mW_{1,\ell}\}_{\ell=1}^{L}$

\For{each adapted weight $\ell$}
    \State Obtain the first-stage subspace projectors
    $\mP_{U,\ell}$ and $\mP_{V,\ell}$
    \For{each calibration batch $\gC_b$}
        \State Compute the task gradient
        $\mG_{\ell,b}$ at
        $\mW_0+\Delta\mW_1$
        \State Form
        $\mG_{\mathrm{res},\ell,b}
        \gets
        \mG_{\ell,b}
        -\mP_{U,\ell}\mG_{\ell,b}\mP_{V,\ell}$
        \State Store its rank-$q$ compressed factors
        and discard the dense gradient
    \EndFor
    \State Aggregate the compressed factors and
    extract the leading $r_2$ right singular vectors
    $\mV_{\ell,r}$
    \State Initialize
    $\mA_{2,\ell}\gets\mV_{\ell,r}^{\top}$,
    $\mB_{2,\ell}\gets\mathbf{0}$
\EndFor

\State Fix Stage 1 and optimize only the Stage-2 factors
using the task loss and orthogonality regularizer

\For{each adapted weight $\ell$}
    \State Merge
    $\mW_{\ell}^{*}
    \gets
    \mW_{0,\ell}
    +\Delta\mW_{1,\ell}
    +\Delta\mW_{2,\ell}$
\EndFor

\State \Return $\{\mW_{\ell}^{*}\}_{\ell=1}^{L}$
\end{algorithmic}
\end{algorithm}

\subsection{Computational Cost}
\label{app:complexity}

For an adapted weight of size
$d_{\mathrm{out}}\times d_{\mathrm{in}}$, each dense
calibration gradient is held only transiently. Storing
rank-$q$ factors for $M$ batches requires
$O\!\left(Mq(d_{\mathrm{out}}+d_{\mathrm{in}})\right)$
numbers per adapted weight. Factorized aggregation uses QR
factorizations and an SVD of a core matrix of size at most
$Mq\times Mq$; it does not explicitly form the dense
aggregate gradient. Calibration is performed only once
between the training stages.

The orthogonality penalty is computed from LoRA factors
rather than dense update matrices. Its per-weight cost is
\begin{equation}
O\!\left(
(d_{\mathrm{out}}+d_{\mathrm{in}})
(r_1^2+r_2^2+r_1r_2)
\right).
\end{equation}
Across the two stages, the LoRA factors contain
$(r_1+r_2)(d_{\mathrm{out}}+d_{\mathrm{in}})$
parameters per adapted weight, excluding any task-specific
output head. Once both updates are merged, inference uses
the same dense weight multiplication as the pretrained model.

\section{Additional Results}
\label{app:additional-results}

\subsection{Commonsense Reasoning}
\label{app:commonsense-results}

\begin{table}[H]
\caption{\textbf{Accuracy (\%) on eight commonsense reasoning benchmarks}
using Llama-3-8B. Avg. is the unweighted mean across tasks. Subscript values
beside ReGRA scores show percentage-point differences from rank-8 LoRA.
The best average is marked in bold.}
\label{tab:commonsense-results}
\centering
\renewcommand{\arraystretch}{1.08}
\setlength{\tabcolsep}{2.2pt}

\definecolor{loraSixteenGray}{RGB}{125,125,125}
\definecolor{loraEightTint}{RGB}{240,240,240}
\definecolor{deltaGray}{RGB}{105,105,105}
\definecolor{regraTint}{RGB}{229, 245, 250}

\def\datasetHeader#1{%
    \makebox[2.8em][c]{\textbf{#1}}%
    \makebox[1.9em][l]{}%
}
\def\alignedScore#1{%
    \makebox[2.8em][r]{#1}%
    \makebox[1.9em][l]{}%
}
\def\scoreCell#1{\alignedScore{#1}}
\def\grayScoreCell#1{\cellcolor{loraEightTint}\alignedScore{#1}}
\def\loraSixteenScore#1{%
    \cellcolor{white}%
    \alignedScore{\textcolor{loraSixteenGray}{#1}}%
}
\def\regraCell#1#2{%
    \cellcolor{regraTint}%
    \makebox[2.8em][r]{#1}%
    \makebox[1.9em][l]{\ensuremath{{}_{\textcolor{deltaGray}{#2}}}}%
}

\resizebox{\textwidth}{!}{%
\begin{tabular}{@{}cl|cccccccc|c@{}}
\toprule
& \textbf{Method}
& \datasetHeader{OBQA}
& \datasetHeader{ARC-C}
& \datasetHeader{Wino.}
& \datasetHeader{PIQA}
& \datasetHeader{SIQA}
& \datasetHeader{ARC-E}
& \datasetHeader{BoolQ}
& \datasetHeader{Hella.}
& \datasetHeader{Avg.} \\
\midrule
\multirow{6}{*}{\rotatebox[origin=c]{90}{Llama-3-8B}}
& \textcolor{loraSixteenGray}{LoRA ($r=16$)}
& \loraSixteenScore{88.40}
& \loraSixteenScore{81.10}
& \loraSixteenScore{87.29}
& \loraSixteenScore{88.74}
& \loraSixteenScore{81.63}
& \loraSixteenScore{93.14}
& \loraSixteenScore{74.19}
& \loraSixteenScore{93.35}
& \loraSixteenScore{85.98} \\
\cmidrule(l){2-11}
& \cellcolor{loraEightTint}LoRA ($r=8$)
& \grayScoreCell{86.80}
& \grayScoreCell{81.31}
& \grayScoreCell{87.61}
& \grayScoreCell{88.57}
& \grayScoreCell{81.12}
& \grayScoreCell{93.10}
& \grayScoreCell{69.93}
& \grayScoreCell{92.73}
& \grayScoreCell{85.15} \\
& DoRA ($r=8$)
& \scoreCell{86.40}
& \scoreCell{81.14}
& \scoreCell{87.21}
& \scoreCell{89.28}
& \scoreCell{81.57}
& \scoreCell{93.43}
& \scoreCell{73.82}
& \scoreCell{92.57}
& \scoreCell{85.68} \\
& PiSSA ($r=8$)
& \scoreCell{81.00}
& \scoreCell{77.90}
& \scoreCell{86.02}
& \scoreCell{88.46}
& \scoreCell{80.80}
& \scoreCell{92.38}
& \scoreCell{73.18}
& \scoreCell{91.55}
& \scoreCell{83.91} \\
& LoRA-One ($r=8$)
& \scoreCell{88.20}
& \scoreCell{81.05}
& \scoreCell{87.37}
& \scoreCell{88.84}
& \scoreCell{81.67}
& \scoreCell{93.22}
& \scoreCell{74.31}
& \scoreCell{92.81}
& \scoreCell{85.93} \\
\cmidrule(l){2-11}
& \cellcolor{regraTint}\textbf{ReGRA ($r=8$)}
& \regraCell{88.20}{+1.40}
& \regraCell{82.00}{+0.69}
& \regraCell{86.90}{-0.71}
& \regraCell{89.34}{+0.77}
& \regraCell{81.93}{+0.81}
& \regraCell{92.30}{-0.80}
& \regraCell{74.22}{+4.29}
& \regraCell{93.07}{+0.34}
& \regraCell{\textbf{86.00}}{+0.85} \\
\bottomrule
\end{tabular}}
\end{table}

Table~\ref{tab:commonsense-results} shows that ReGRA achieves the
highest eight-task average on Llama-3-8B, improving upon rank-8 LoRA
by 0.85 percentage points. The largest task-level gains occur on BoolQ
and OBQA, while Wino. and ARC-E decline. Increasing standard LoRA to
rank 16 also improves the average and nearly matches ReGRA (85.98\%
versus 86.00\%). These results extend the average gain over rank-8
LoRA to commonsense reasoning.

\subsection{Error-Level Analysis of Stage-2 Components}
\label{app:stage2-error-analysis}
\subsubsection{Unique Error Corrections}
\label{app:unique-error-corrections}
\begin{figure}[t]
\centering
\includegraphics[width=0.92\linewidth]{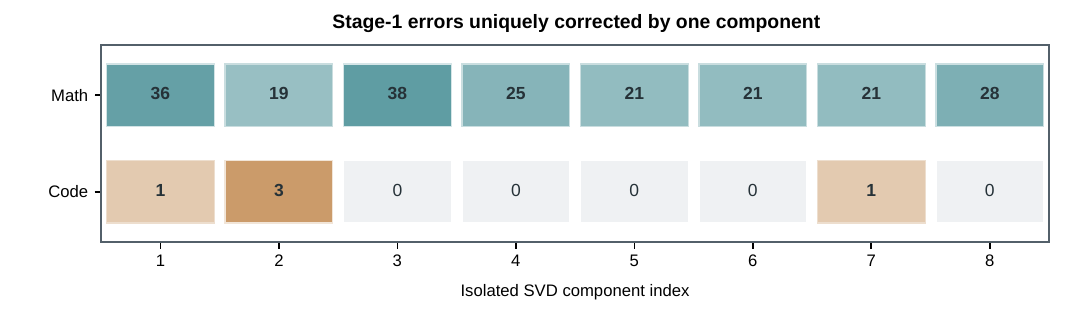}
\caption{\textbf{Stage 1 errors uniquely corrected by isolated
Stage-2 components.} Each cell counts cases answered incorrectly by
Stage 1, corrected by the indicated component, and not corrected by
any of the other seven isolated components. Math counts task--example
pairs across six benchmarks; Code counts MBPP+ problems.}
\label{fig:stage2-component-unique-corrections}
\end{figure}

Figure~\ref{fig:stage2-component-unique-corrections} compares the
Stage-1 errors corrected by each isolated Stage-2 component. On Math,
all eight components uniquely corrected errors, with 19--38
task--example pairs per component. On MBPP+, only components 1, 2,
and 7 had unique corrections, with 1, 3, and 1 problems, respectively.
The isolated components therefore showed distinct correction patterns
on Math, while this distinction was less evident on MBPP+.

\subsubsection{Overlap with Full-Model Gains}
\label{app:full-model-gains-overlap}
\begin{figure}[H]
\centering
\includegraphics[width=0.92\linewidth]{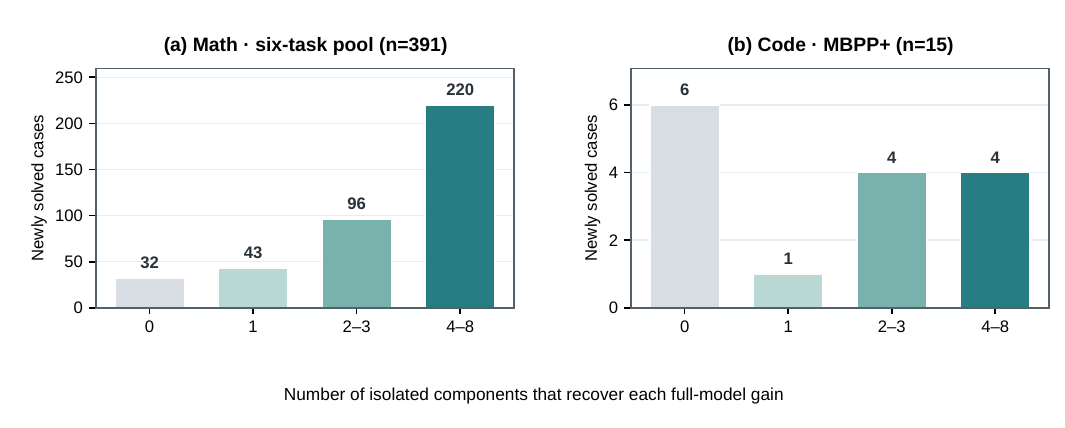}
\caption{\textbf{Coverage of the full Stage-2 model's newly solved
cases.} For each case answered incorrectly by Stage 1 and correctly
by the full Stage-2 model, the bars show how many isolated components
also answer it correctly. Math includes 391 task--example pairs; Code
includes 15 MBPP+ problems.}
\label{fig:stage2-component-full-gain-coverage}
\end{figure}

Figure~\ref{fig:stage2-component-full-gain-coverage} examines cases
newly solved by the full Stage-2 model. At least one isolated component
also solved 359 of 391 Math cases and 9 of 15 MBPP+ problems. The
remaining 32 Math cases and 6 MBPP+ problems were not solved by any
isolated component. These results show that isolated components recover
many full-model gains, while some gains appear only with the combined
update. The analysis does not measure each component's marginal
contribution within that update.

\end{document}